\documentclass[letterpaper, 10 pt, conference]{ieeeconf}  

\IEEEoverridecommandlockouts                              

\usepackage{graphicx} 
\usepackage{amsmath} 
\usepackage{amssymb}  
\usepackage[unicode=true, colorlinks=true, linkcolor=blue]{hyperref}
\usepackage{color}
\usepackage[T1]{fontenc}

\DeclareMathOperator*{\argmax}{argmax}
\newcommand{\etal}{\textit{et al}. }
\newcommand{\ie}{\textit{i}.\textit{e}. }
\newcommand{\eg}{\textit{e}.\textit{g}. }
\newcommand{\mypara}{\par\vspace*{0.5mm}\noindent\textbf}

\title{\LARGE \bf
Learning Synergies between Pushing and Grasping\\ with Self-supervised Deep Reinforcement Learning
\vspace{-1mm}
}

\author{ 
Andy Zeng$^{1,2}$, 
Shuran Song$^{1,2}$, 
Stefan Welker$^{2}$,
Johnny Lee$^{2}$,
Alberto Rodriguez$^{3}$, 
Thomas Funkhouser$^{1,2}$ 
\vspace{1mm} \\ 
$^{1}$Princeton University\quad\quad
$^{2}$Google\quad\quad
$^{3}$Massachusetts Institute of Technology
\vspace{2mm} \\
\href{http://vpg.cs.princeton.edu/}{http://vpg.cs.princeton.edu}
\vspace{-3mm}
\thanks{The authors would like to thank NSF (VEC 1539014/1539099), Google, Amazon, Intel, NVIDIA, ABB Robotics, and Mathworks for hardware, technical, and financial support.}
}

\begin{document}
\tracingall

\maketitle
\thispagestyle{empty}
\pagestyle{empty}

\begin{abstract}

Skilled robotic manipulation benefits from complex synergies between non-prehensile (\eg pushing) and prehensile (\eg grasping) actions: pushing can help rearrange cluttered objects to make space for arms and fingers; likewise, grasping can help displace objects to make pushing movements more precise and collision-free.
In this work, we demonstrate that it is possible to discover and learn these synergies from scratch through model-free deep reinforcement learning. Our method involves training two fully convolutional networks that map from visual observations to actions: one infers the utility of pushes for a dense pixel-wise sampling of end effector orientations and locations, while the other does the same for grasping. 
Both networks are trained jointly in a Q-learning framework and are entirely self-supervised by trial and error, where rewards are provided from successful grasps. In this way, our policy learns pushing motions that enable future grasps, while learning grasps that can leverage past pushes. 
During picking experiments in both simulation and real-world scenarios, we find that our system quickly learns complex behaviors amid challenging cases of clutter, and achieves better grasping success rates and picking efficiencies than baseline alternatives after only a few hours of training. We further demonstrate that our method is capable of generalizing to novel objects. Qualitative results (videos), code, pre-trained models, and simulation environments are available at \href{http://vpg.cs.princeton.edu/}{http://vpg.cs.princeton.edu}

\end{abstract}

\section{Introduction}

\begin{figure}[t]
\centering
  \vspace{2mm}
  \includegraphics[width=\linewidth]{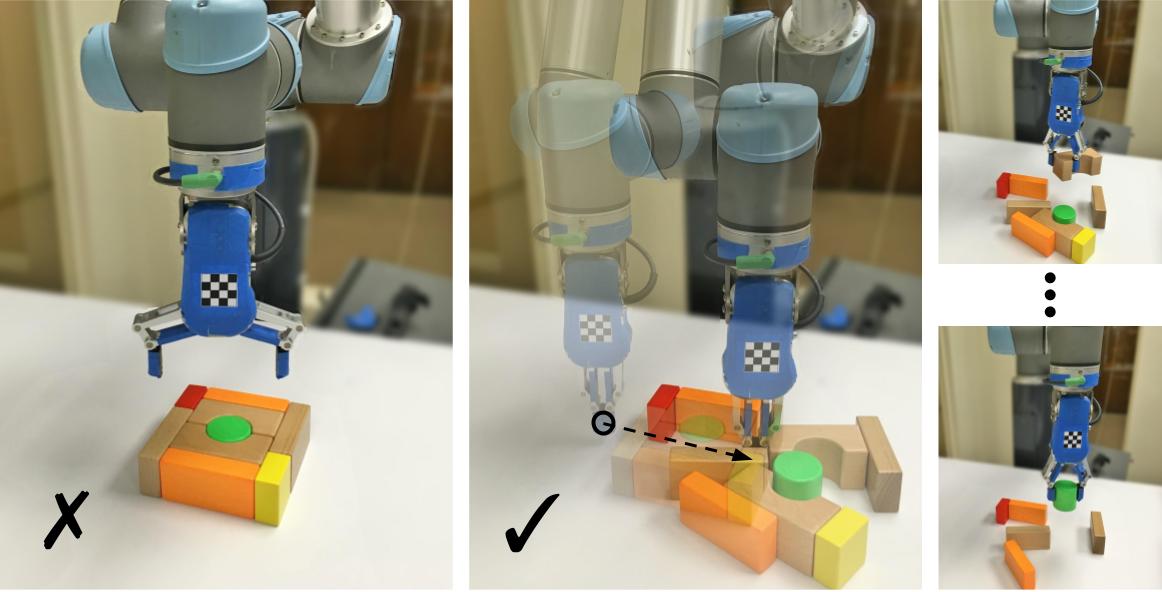}
  \caption{\textbf{Example configuration of tightly packed blocks} reflecting the kind of clutter that commonly appears in real-world scenarios (\eg with stacks of books, boxes, etc.), which remains challenging for grasp-only manipulation policies.
  Our model-free system is able to plan pushing motions that can isolate these objects from each other, making them easier to grasp; improving the overall stability and efficiency of picking.
  }
  \vspace{-3mm}
  \label{fig:teaser}
\end{figure}

Skilled manipulation benefits from the synergies between non-prehensile (\eg pushing) and prehensile (\eg grasping) actions:
pushing can help rearrange cluttered objects to make space for arms and fingers (see Fig. \ref{fig:teaser}); likewise, grasping can help displace objects to make pushing movements more precise and collision-free.

Although considerable research has been devoted to both push and grasp planning, they have been predominantly studied in isolation. 
Combining pushing and grasping policies for sequential manipulation is a relatively unexplored problem. Pushing is traditionally studied for the task of precisely controlling the pose of an object. However, in many of the synergies between pushing and grasping, pushing plays a loosely defined role, \eg separating two objects, making space in a particular area, or breaking up a cluster of objects. These goals are difficult to define or reward for model-based~\cite{mason1986mechanics,goyal1991planar,hogan2016feedback} or data-driven~\cite{bauza2017probabilistic,finn2017deep,clavera2017policy} approaches.


Many recent successful approaches to learning grasping policies, maximize affordance metrics learned from experience~\cite{pinto2016supersizing,zeng2017robotic} or induced by grasp stability metrics~\cite{gualtieri2016high,mahler2017dex}. However, it remains unclear how to plan sequences of actions that combine grasps and pushes, each learned in isolation.
%
%
While hard-coded heuristics for supervising push-grasping policies have been successfully developed by exploiting domain-specific knowledge~\cite{dogar2012planning}, they limit the types of synergistic behaviors between pushing and grasping that can be performed.

In this work, we propose to discover and learn synergies between pushing and grasping from experience through model-free deep reinforcement learning (in particular, Q-learning).
The key aspects of our system are:
\begin{itemize}
    \item We learn joint pushing and grasping policies through self-supervised trial and error. Pushing actions are useful only if, in time, enable grasping. This is in contrast to prior approaches that define heuristics or hard-coded objectives for pushing motions.
    \item We train our policies end-to-end with a deep network that takes in visual observations and outputs expected return (\ie in the form of Q values) for potential pushing and grasping actions. The joint policy then chooses the action with the highest Q value -- \ie, the one that maximizes the expected success of current/future grasps. This is in contrast to explicitly perceiving individual objects and planning actions on them based on hand-designed features~\cite{boularias2015learning}. 
\end{itemize}
This formulation enables our system to execute complex sequential manipulations (with pushing and grasping) of objects in unstructured picking scenarios and generalizes to novel objects (unseen in training).

Training deep end-to-end policies (\eg from image pixels to joint torques) with reinforcement learning on physical systems can be expensive and time-consuming due to their prohibitively high sample complexity \cite{levine2016end,popov2017data,rajeswaran2017learning}. 
To make training tractable on a real robot, we simplify the action space to a set of end-effector-driven motion primitives.  We formulate the task as a pixel-wise labeling problem: where each image pixel -- and image orientation -- corresponds to a specific robot motion primitive (pushing or grasping) executed on the 3D location of that pixel in the scene. For pushing, this location represents the starting position of the pushing motion; for grasping, the middle position between the two fingers during parallel-jaw grasping. We train a fully convolutional network (FCN) to take an image of the scene as input, and infer dense pixel-wise predictions of future expected reward values for all pixels -- and thereby all robot motion primitives executed for all visible surfaces in the scene.
This pixel-wise parameterization of robot primitive actions, which we refer to as \textbf{fully convolutional action-value functions} \cite{zeng2017robotic}, enables us to train effective pushing and grasping policies on a single robot arm in less than a few hours of robot time.

The main contribution of this paper is a new perspective to bridging data-driven prehensile and non-prehensile manipulation. We demonstrate that it is 
possible to train end-to-end deep networks to capture complementary pushing and grasping policies that benefit from each other through experience.
We provide several experiments and ablation studies in both simulated and real settings to evaluate the key components of our system. Our results show that the pushing policies enlarge the set of scenarios in which grasping succeeds, and that both policies working in tandem produce complex interactions with objects (beyond our expectations) that support more efficient picking (\eg pushing multiple blocks at a time, separating two objects, breaking up a cluster of objects through a chain of reactions that improves grasping). We provide additional qualitative results (video recordings of our robot in action), code, pre-trained models, and simulation environments at \href{http://vpg.cs.princeton.edu/}{http://vpg.cs.princeton.edu}

\section{Related Work}

Our work lies at the intersection of robotic manipulation, computer vision, and machine learning. We briefly review the related work in these domains.

\mypara{Non-prehensile manipulation.} Planning non-prehensile motions, such as pushing, is a fundamental problem that dates back to the early days of robotic manipulation. The literature in this area is vast, emerging early from classical solutions that explicitly model the dynamics of pushing with frictional forces \cite{mason1986mechanics,goyal1991planar}. While inspiring, many of these methods rely on modeling assumptions that do not hold in practice \cite{yu2016more,bauza2017probabilistic}. For example, non-uniform friction distributions across object surfaces and the variability of friction are only some of the factors that can lead to erroneous predictions of friction-modeling pushing solutions in real-world settings. While recent methods have explored data-driven algorithms for learning the dynamics of pushing \cite{salganicoff1993vision,mericcli2015push,zhou2016convex}, many of these works have largely focused on the execution of stable pushes for one object at a time. Modeling the larger-scale consequences of pushing in the face of severe clutter and friction variation continues to be a complex problem; effectively using these models to discover optimal policies in real settings -- even more so. 


\mypara{Grasping.} Grasping too, has been well studied in the domain of model-based reasoning; from modeling contact forces and their resistance to external wrenches \cite{prattichizzo2008grasping,weisz2012pose}, to characterizing grasps by their ability to constrain object mobility \cite{rodriguez2012caging}. A common approach to deploying these methods in real systems involves pre-computing grasps from a database of known 3D object models \cite{goldfeder2009columbia}, and indexing them at run-time with point cloud registration for object pose estimation \cite{zeng2017multi,zeng20173dmatch}. These methods, however, typically assume knowledge of object shapes, poses, dynamics, and contact points -- information which is rarely known for novel objects in unstructured environments.

More recent data-driven methods explore the prospects of training model-agnostic deep grasping policies \cite{redmon2015real,pinto2016supersizing,pinto2017learning,gualtieri2016high,mahler2017dex,zeng2017robotic} that detect grasps by exploiting learned visual features, and without explicitly using object specific knowledge (\ie shape, pose, dynamics).
Pinto \etal \cite{pinto2017learning} improve the performance of these deep policies by using models pre-trained on auxiliary tasks such as poking. Zeng \etal \cite{zeng2017robotic} demonstrate that using FCNs to efficiently model these policies with affordances can drastically improve run-times. Analogous to these methods, our data-driven framework is model-agnostic, but with the addition of improving the performance of grasping by incorporating non-prehensile actions like pushing.


\begin{figure*}[t]
\centering
  \includegraphics[width=\linewidth]{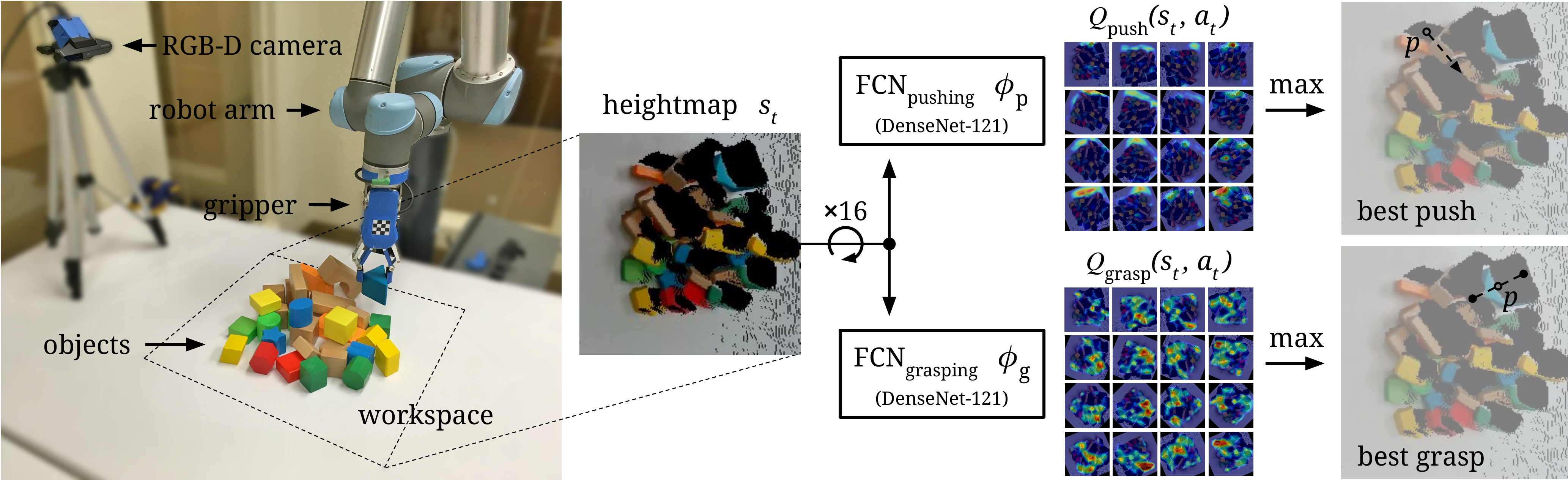}
  \caption{{\bf Overview} of our system and Q-learning formulation. Our robot arm operates over a workspace observed by a statically mounted RGB-D camera. Visual 3D data is re-projected onto an orthographic RGB-D heightmap, which serves as a representation of the current state $s_t$. The heightmaps are then fed into two FCNs - one $\phi_\mathrm{p}$ inferring pixel-wise Q values (visualized with heat maps) for pushing to the right of the heightmap and another $\phi_\mathrm{g}$ for horizontal grasping over the heightmap. Each pixel represents a different location on which to execute the primitive. This is repeated for 16 different rotations of the heightmap to account for various pushing and grasping angles. These FCNs jointly define our deep Q function and are trained simultaneously.}
  \label{fig:method}
  \vspace{-3mm}
\end{figure*}

\mypara{Pushing with grasping.} Combining both non-prehensile and prehensile manipulation policies is interesting, albeit an area of research that has been much less explored. The seminal work of Dogar \etal \cite{dogar2012planning} presents a robust planning framework for push-grasping (non-prehensile motions baked within grasping primitives) to reduce grasp uncertainty as well as an additional motion primitive -- sweeping -- to move around obstacles in clutter. The policies in their framework, however, remain largely handcrafted. In contrast, our method is data-driven and learned online by self-supervision.

Other methods \cite{omrvcen2009autonomous,clavera2017policy} explore the model-free planning of pushing motions to move objects to target positions that are more favorable for pre-designed grasping algorithms -- the behaviors of which are typically handcrafted, fixed, and well-known in advance. This knowledge is primarily used to define concrete goals (\eg target positions) that can aid in the design or training of pushing policies. However, trying to define similar goals for data-driven model-agnostic grasping policies (where optimal behaviors emerge from experience) become less clear, as these policies are constantly learning, changing, and adapting behaviors over time with more data.

More closely related to our work is that of Boularias \etal \cite{boularias2015learning}, which explores the use of reinforcement learning for training control policies to select among push and grasp proposals represented by hand-crafted features. They propose a pipeline that first segments images into objects, proposes pushing and grasping actions, extracts hand-tuned features for each action, then executes the action with highest expected reward. While inspiring, their method models perception and control policies separately (not end-to-end); it relies on model-based simulation to predict the motion of pushed objects and to infer its benefits for future grasping (those predictions are the two ``features'' provided to the pushing policy); it is tuned to work mainly for convex objects, and demonstrated on only one scenario with only two objects (a cylinder next to a box). In contrast, we train perception and control policies with end-to-end deep networks; we make no assumptions about the shapes or dynamics of objects (model-free), and we demonstrate that our formulation works not only for a variety of test cases with numerous objects (up to 30+), but also that it is capable of quickly generalizing to novel objects and scenarios. To the best of our knowledge, our work is the first model-free system to perform reinforcement learning of complementary pushing and grasping policies with deep networks that operate end-to-end from visual observations to actions.

\section{Problem Formulation}

We formulate the task of pushing-for-grasping as a Markov decision process: at any given state $s_t$ at time $t$, the agent (\ie robot) chooses and executes an action $a_t$ according to a policy $\pi(s_t)$, then transitions to a new state $s_{t+1}$ and receives an immediate corresponding reward $R_{a_t}(s_t,s_{t+1})$. The goal of our robotic reinforcement learning problem is to find an optimal policy $\pi^*$ that maximizes the expected sum of future rewards, given by $R_t=\sum_{i=t}^{\infty}\gamma R_{a_i}(s_i,s_{i+1})$, \ie $\gamma$-discounted sum over an infinite-horizon of future returns from time $t$ to $\infty$.

In this work, we investigate the use of off-policy Q-learning to train a greedy deterministic policy $\pi(s_t)$ that chooses actions by maximizing the action-value function (\ie Q-function) $Q_\pi(s_t,a_t)$, which measures the expected reward of taking action $a_t$ in state $s_t$ at time $t$. Formally, our learning objective is to iteratively minimize the temporal difference error $\delta_t$ of $Q_\pi(s_t,a_t)$ to a fixed target value $y_t$: 
$$\delta_t = |Q(s_t,a_t) - y_t|$$
$$y_t=R_{a_t}(s_t,s_{t+1})+\gamma\,Q(s_{t+1},\argmax_{a'}(Q(s_{t+1},a')))$$
\noindent where $a'$ is the set of all available actions. 

\section{Method}
\label{sec:method}

This section provides details of our Q-learning formulation, network architectures, and training protocols.

\subsection{State Representations}

We model each state $s_t$ as an RGB-D heightmap image representation of the scene at time $t$. To compute this heightmap, we capture RGB-D images from a fixed-mount camera, project the data onto a 3D point cloud,
and orthographically back-project upwards in the gravity direction to construct a heightmap image representation with both color (RGB) and height-from-bottom (D) channels (see Fig. \ref{fig:method}). The edges of the heightmaps are predefined with respect to the boundaries of the agent's workspace for picking. In our experiments, this area covers a $0.448^2$m tabletop surface. Since our heightmaps have a pixel resolution of $224\times224$, each pixel spatially represents a $2^2$mm vertical column of 3D space in the agent's workspace.

\subsection{Primitive Actions}

We parameterize each action $a_t$ as a motion primitive behavior $\psi$ (\eg pushing or grasping) executed at the 3D location $q$ projected from a pixel $p$ of the heightmap image representation of the state $s_t$: 
$$a = (\psi,q)\: |\: \psi \in \{\mathrm{push},\mathrm{grasp}\}, q \twoheadrightarrow p \in s_t$$ 
Our motion primitive behaviors are defined as follows:
\mypara{Pushing:} $q$ denotes the starting position of a 10cm push in one of $k=16$ directions. The trajectory of the push is straight.  It is physically executed in our experiments using the tip of a closed two-finger gripper.
\mypara{Grasping:} $q$ denotes the middle position of a top-down parallel-jaw grasp in one of $k=16$ orientations. During a grasp attempt, both fingers attempt to move 3cm below $q$ (in the gravity direction) before closing the fingers. In both primitives, robot arm motion planning is automatically executed with stable, collision-free IK solves \cite{diankov_thesis}.

\subsection{Learning Fully Convolutional Action-Value Functions}


We extend vanilla deep Q-networks (DQN) \cite{mnih2015human} by modeling our Q-function as two feed-forward fully convolutional networks (FCNs) \cite{long2015fully} $\phi_\mathrm{p}$ and $\phi_\mathrm{g}$; one for each motion primitive behavior (pushing and grasping respectively). Each individual FCN $\phi_\psi$ takes as input the heightmap image representation of the state $s_t$ and outputs a dense pixel-wise map of Q values with the same image size and resolution as that of $s_t$, where each individual Q value prediction at a pixel $p$ represents the future expected reward of executing primitive $\psi$ at 3D location $q$ where $q \twoheadrightarrow p \in s_t$. Note that this formulation is a direct amalgamation of Q-learning with visual affordance-based manipulation \cite{zeng2017robotic}.

Both FCNs $\phi_\mathrm{p}$ and $\phi_\mathrm{g}$ share the same network architecture: two parallel 121-layer DenseNet \cite{huang2017densely} pre-trained on ImageNet \cite{deng2009imagenet}, followed by channel-wise concatenation and 2 additional $1\times1$ convolutional layers interleaved with nonlinear activation functions (ReLU) \cite{nair2010rectified} and spatial batch normalization \cite{ioffe2015batch}, then bilinearly upsampled. One DenseNet tower takes as input the color channels (RGB) of the heightmap, while the other takes as input the channel-wise cloned depth channel (DDD) (normalized by subtracting mean and dividing standard deviation) of the heightmap.

To simplify learning oriented motion primitives for pushing and grasping, we account for different orientations by rotating the input heightmap $s_t$ into $k=16$ orientations (different multiples of $22.5^\circ$) and then consider only horizontal pushes (to the right) and grasps in the rotated heightmaps.  Thus, the input to each FCN $\phi_\psi$ is $k=16$ rotated heightmaps, and the total output is 32 pixel-wise maps of Q values (16 for pushes in different directions, and 16 for grasps at different orientations). The action that maximizes the Q-function is the primitive and pixel with the highest Q value across all 32 pixel-wise maps: $\argmax_{a'_{t}}(Q(s_{t},a'_{t})) = \argmax_{(\psi,p)}(\phi_\mathrm{p}(s_t),\phi_\mathrm{g}(s_t))$. 

Our pixel-wise parameterization of both state and action spaces enables the use of FCNs as Q-function approximators, which provides several advantages. First, the Q value prediction for each action now has an explicit notion of spatial locality with respect to other actions, as well as to the input observation of the state (\eg with receptive fields). Second, FCNs are efficient for pixel-wise computations. Each forward pass of our network architecture $\phi_\psi$ takes on average 75ms to execute, which enables computing Q values for all 1,605,632 (\ie $224\times224\times32$) possible actions within 2.5 seconds. Finally, our FCN models can converge with less training data since the parameterization of end effector locations (pixel-wise sampling) and orientations (by rotating $s_t$) enables convolutional features to be shared across locations and orientations (\ie equivariance to translation and rotation).

Additional extensions to deep networks for Q-function estimation such as double Q-learning \cite{van2016deep}, and duelling networks \cite{wang2015dueling}, have the potential to improve performance but are not the focus of this work.


\subsection{Rewards}
\label{sec:rewards}

Our reward scheme for reinforcement learning is simple. We assign $R_{\mathrm{g}}(s_t,s_{t+1})=1$ if a grasp is successful (computed by thresholding on the antipodal distances between gripper fingers after a grasp attempt) and $R_{\mathrm{p}}(s_t,s_{t+1})=0.5$ for pushes that make detectable changes to the environment (where changes are detected if the sum of differences between heightmaps exceeds some threshold $\tau$, \ie $\sum (s_{t+1}-s_t) > \tau$).  Note that the intrinsic reward $R_{\mathrm{p}}(s_t,s_{t+1})$ does not explicitly consider whether a push enables future grasps. Rather, it simply encourages the system to make pushes that cause change. The synergy between pushing and grasping is learned mainly through reinforcement (see experiments in Sec. \ref{sec:simulation-results}).

\subsection{Training details.}
\label{sec:training-details}

Our Q-learning FCNs are trained at each iteration $i$ using the Huber loss function:
$$\mathcal{L}_i=\left\{\begin{array}{ll} \frac{1}{2}(Q^{\theta_i}(s_i,a_i) - y_i^{\theta_i^-})^2, \mathrm{for}\:|Q^{\theta_i}(s_i,a_i) - y_i^{\theta_i^-}| < 1,\\
                  |Q^{\theta_i}(s_i,a_i) - y_i^{\theta_i^-}| - \frac{1}{2},\:\mathrm{otherwise.}\end{array}\right.$$
\noindent where $\theta_i$ are the parameters of the neural network at iteration $i$, and the target network parameters $\theta_i^-$ are held fixed between individual updates. We pass gradients only through the single pixel $p$ and network $\phi_\mathrm{\psi}$ from which the value predictions of the executed action $a_i$ was computed. All other pixels at iteration $i$ backpropagate with 0 loss.

We train our FCNs $\phi_\psi$ by stochastic gradient descent with momentum, using fixed learning rates of $10^{-4}$, momentum of 0.9, and weight decay $2^{-5}$. Our models are trained in PyTorch with an NVIDIA Titan X on an Intel Xeon CPU E5-2699 v3 clocked at 2.30GHz. We train with prioritized experience replay \cite{schaul2015prioritized} using stochastic rank-based prioritization, approximated with a power-law distribution. Our exploration strategy is $\epsilon$-greedy, with $\epsilon$ initialized at 0.5 then annealed over training to 0.1. Our future discount $\gamma$ is constant at 0.5.

In our experiments (Sec. \ref{sec:experiments}), we train all of our models by self-supervision with the same procedure: $n$ objects (\ie toy blocks) are randomly selected and dropped into the $0.448^2$m workspace in front of the robot. The robot then automatically performs data collection by trial and error, until the workspace is void of objects, at which point $n$ objects are again randomly dropped into the workspace. In simulation $n=10$, while in real-world settings $n=30$.

\subsection{Testing details.}
\label{sec:testing-details}

Since our policy is greedy deterministic during test time, it is possible for it to get stuck repeatedly executing the same action while the state representation (and thus value estimates) remain the same as no change is made to the environment. Naively weighting actions based on visit counts can also be inefficient due our pixel-wise parameterization of the action space. Hence to alleviate this issue, during testing we prescribe a small learning rate to the network at $10^{-5}$ and continue backpropagating gradients through the network after each executed action. For the purposes of evaluation, network weights are reset to their original state (after training and before testing) prior to each new experiment test run -- indicated by when all objects in the workspace have been successfully grasped (\ie completion) or when the number of consecutively executed actions for which there is no change to the environment exceeds 10.

\section{Experiments}
\label{sec:experiments}

We executed a series of experiments to test the proposed approach, which we call \textbf{Visual Pushing for Grasping (VPG)}.  The goals of the experiments are three-fold: 1) to investigate whether the addition of pushing as a motion primitive can enlarge the set of scenarios in which objects can successfully be grasped (\ie does pushing help grasping), 2) to test whether it is feasible to train pushing policies with supervision mainly from the future expected success of another grasping policy trained simultaneously, and 3) to demonstrate that our formulation is capable of training effective, non-trivial pushing-for-grasping policies directly from visual observations on a real system.  

\subsection{Baseline Methods}
\label{sec:baseline-methods}

To address these goals, we compare the picking performance of VPG to the following baseline approaches:

\mypara{Reactive Grasping-only Policy (Grasping-only)} is a grasping policy that uses the same pixel-wise state and action space formulation as our proposed method described in Section \ref{sec:method}, but uses a single FCN supervised with binary classification (from trial and error) to infer pixel-wise affordance values between 0 and 1 for grasping only. This baseline is a greedy deterministic policy that follows the action which maximizes the immediate grasping affordance value at every time step $t$. This baseline is analogous to a self-supervised version of a state-of-the-art top-down parallel-jaw grasping algorithm \cite{zeng2017robotic}. For a fair comparison, we extend that method using DenseNet \cite{huang2017densely} pre-trained on ImageNet \cite{deng2009imagenet}.

\mypara{Reactive Pushing and Grasping Policy (P+G Reactive)} is an augmented version of the previous baseline, but with an additional FCN to infer pixel-wise affordance values between 0 and 1 for pushing. Both networks are trained with binary classification from self-supervised trial and error, where pushing is explicitly supervised with a binary value from change detection (as described in Section \ref{sec:rewards}). Change detection is the simplest form of direct supervision for pushing, but requires higher values of $\epsilon$ for the exploration strategy to maintain stable training. This policy follows the action which maximizes the immediate affordance value (which can come from either the pushing or grasping FCNs).

Both aforementioned baselines are reactive as they do not plan long-horizon strategies, but instead greedily choose actions based on affordances computed from the current state $s_t$. Our training optimization parameters for these baselines are kept the same as that of VPG.

\subsection{Evaluation Metrics}
\label{sec:test-protocol}

We test the methods by executing a series of tests in which the system must pick and remove objects from a table with novel arrangements of objects (as described in Sec. \ref{sec:testing-details}).  

For each test, we execute $n$ runs ($n \in {\sim10,30}$) and then evaluate performance with 3 metrics: 1) the average \% completion rate over the $n$ test runs, which measures the ability of the policy to finish the task by picking up all objects without failing consecutively for more than 10 attempts, 2) the average \% grasp success rate per completion, and 3) the \% action efficiency (defined as $\frac{\text{\# objects in test}}{\text{\# actions before completion}}$), which describes how succinctly the policy is capable of finishing the task.  Note that grasp success rate is equivalent to action efficiency for grasping-only policies. For all of these metrics, higher is better. 

We run experiments on both simulated and real-world platforms. While our main objective is to demonstrate effective VPG policies on a real robot, we also run experiments in simulation to provide controlled environments for fair evaluation between methods and for ablation studies. In experiments on both platforms, we run tests with objects placed in both random and challenging arrangements.  

\subsection{Simulation Experiments}
\label{sec:simulation-results}

\begin{figure}[t]
\centering
  \includegraphics[width=\linewidth]{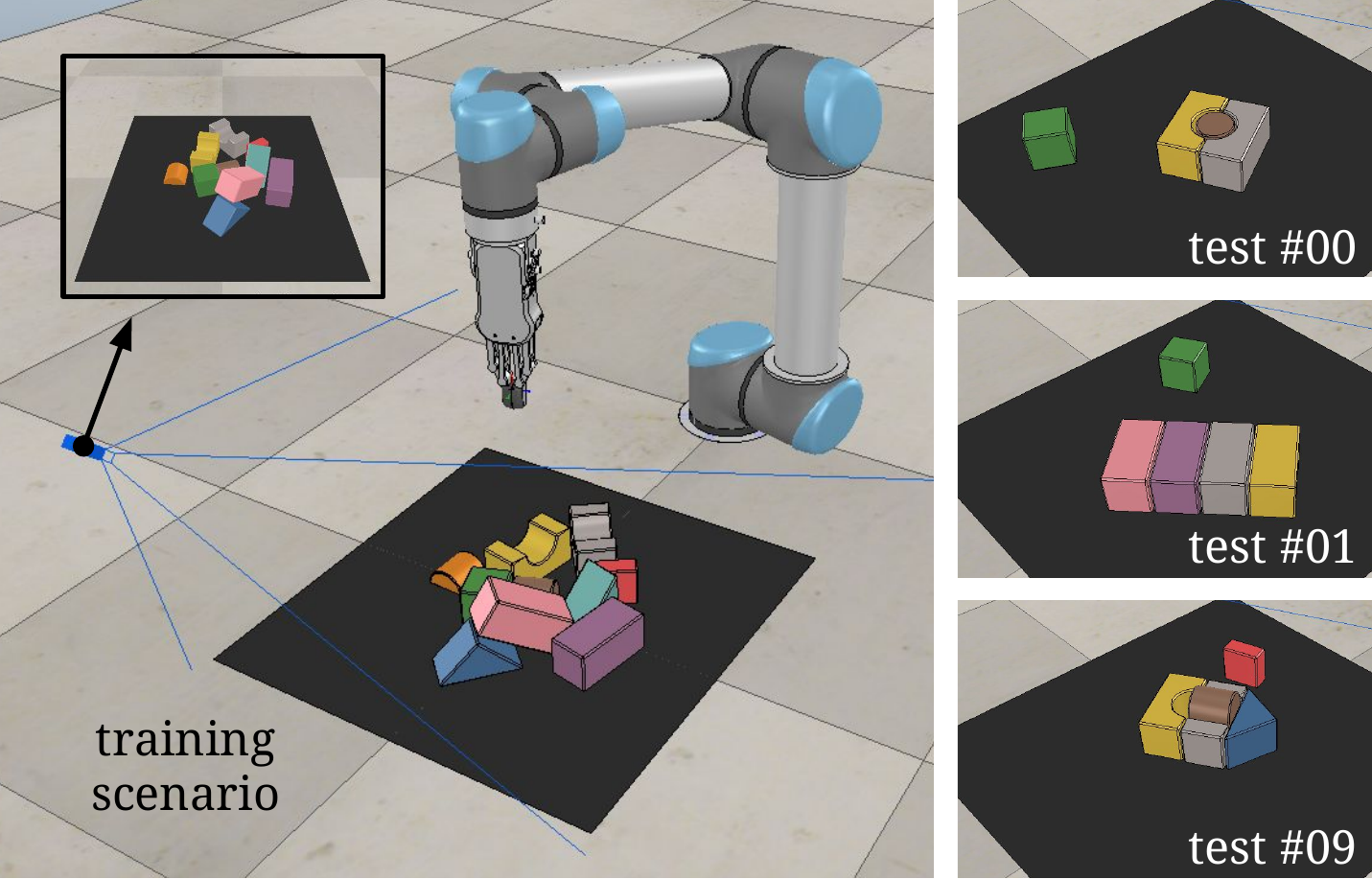}
  \caption{{\bf Simulation environment.} Policies are trained in scenarios with random arrangements of 10 objects (left), then evaluated in scenarios with varying degrees of clutter (10 objects, 30 objects, or challenging object arrangements).  In the most challenging scenarios, adversarial clutter was manually engineered to reflect challenging real-world picking scenarios (\eg tightly packed boxes, as shown on the right).}
  \label{fig:sim-setup}
  \vspace{-3mm}
\end{figure}

Our simulation setup uses a UR5 robot arm with an RG2 gripper in V-REP \cite{rohmer2013vrep} (illustrated in Fig. \ref{fig:sim-setup}) with Bullet Physics 2.83 for dynamics and V-REP's internal inverse kinematics module for robot motion planning.  Each test run in simulation was run $n=30$ times.  The objects used in these simulations include 9 different 3D toy blocks, the shapes and colors of which are randomly chosen during experiments. Most dynamics parameters are kept default except friction coefficients, which have been modified to achieve synthetic object interaction behaviors as similar as possible to that of the real-world. We did not perform any tuning of random seeds for the simulated physics in our experiments. We also simulate a statically mounted perspective 3D camera in the environment, from which perception data is captured. RGB-D images of resolution $640\times480$ are rendered with OpenGL from the camera, without any noise models for depth or color.

\mypara{Comparisons to Baselines.}
Our first experiment compares VPG to the two baseline methods in a simulation where 30 objects are randomly dropped onto a table. This scenario is similar to the training scenario, except it has 30 objects rather than 10, thus testing the generalization of policies to more cluttered scenarios.  
Results are shown in Table \ref{table:sim-test-random}.   
We see that VPG outperforms both baseline methods across all metrics. It is interesting to note that P+G reactive performs poorly in terms of completion rates and action efficiency. This likely due to its tendency (in the face of clutter) to continually push objects around until they are forced out of the workspace as grasping affordances remain low. 

\begin{table}[t]
  \centering
  \setlength{\tabcolsep}{4.0 pt}
  \renewcommand{\arraystretch}{1.1}
  \caption{Simulation Results on Random Arrangements (Mean \%)}
  \begin{tabular}{c|c|c|c}
    \hline
    Method & Completion & Grasp Success & Action Efficiency \\\hline
    Grasping-only \cite{zeng2017robotic} & 90.9 & 55.8 & 55.8 \\
    P+G Reactive & 54.5 & 59.4 & 47.7 \\
    VPG &\bf 100.0 &\bf 67.7 &\bf 60.9 \\\hline
    
  \end{tabular}
  \label{table:sim-test-random}
  \vspace{-3mm}
\end{table}

\mypara{Challenging Arrangements.}
We also compare VPG in simulation to the baseline methods on 11 challenging test cases with adversarial clutter. Each test case consists of a configuration of 3 - 6 objects placed in the workspace in front of the robot, 3 configurations of which are shown in Fig. \ref{fig:sim-setup}. These configurations are manually engineered to reflect challenging picking scenarios, and remain exclusive from the training procedure (described in Sec. \ref{sec:training-details}). Across many of these test cases, objects are laid closely side by side, in positions and orientations that even an optimal grasping policy would have trouble successfully picking up any of the objects without de-cluttering first. As a sanity check, a single isolated object is additionally placed in the workspace separate from the configuration. This is just to ensure that all policies have been sufficiently trained prior to the benchmark (\ie a policy is not ready if fails to grasp the isolated object).

\begin{table}[h]
  \centering
  \setlength{\tabcolsep}{4.0 pt}
  \renewcommand{\arraystretch}{1.1}
  \caption{Simulation Results on Challenging Arrangements (Mean \%)}
  \begin{tabular}{c|c|c|c}
    \hline
    Method & Completion & Grasp Success & Action Efficiency \\\hline
    Grasping-only \cite{zeng2017robotic} & 40.6 & 51.7 & 51.7 \\
    P+G Reactive & 48.2 & 59.0 & 46.4 \\
    VPG &\bf 82.7 &\bf 77.2 &\bf 60.1 \\\hline

  \end{tabular}
  \label{table:sim-hard}
\end{table}

Results are shown in Table \ref{table:sim-hard}.
From the completion results, we observe that the addition of pushing enlarges the set of the scenarios for which successful grasping can be performed. Across the collection of test cases, the grasping-only policy frequently struggles to complete the picking task (with a 0\% completion rate for 5 out of the 11 test cases). We observe this be particularly true in scenarios where large cuboids are laid closely side-by-side (Fig. \ref{fig:sim-setup}).  Even when the policy does successfully complete the task, the average grasp success rates remain relatively low at ~50-60\%. 

Upon the addition of pushing as an additional action primitive in the P+G reactive policy, we immediately see an increase in picking completion rates and there are no longer cases in which the policy completely fails with a 0\% completion rate.  While the P+G reactive policy achieves higher completion and grasp success rates than grasping-only, the average action efficiency is lower. This suggests that the policy executes a large number of pushes, many of which are not succinct and may not actually help grasping. This is expected, since P+G reactive uses binary supervision from change detection for pushing -- pushing motions are not directly supervised by how well they help grasping.

By enabling joint planning of pushing and grasping with VPG, we observe substantially higher completion and grasp success rates (with a 100\% completion rate for 5 of the 11 test cases). The higher action efficiency also indicates that the pushes are now more succinct in how they help grasping.



\begin{figure}[t]
  \centering
  \includegraphics[width=0.9\linewidth]{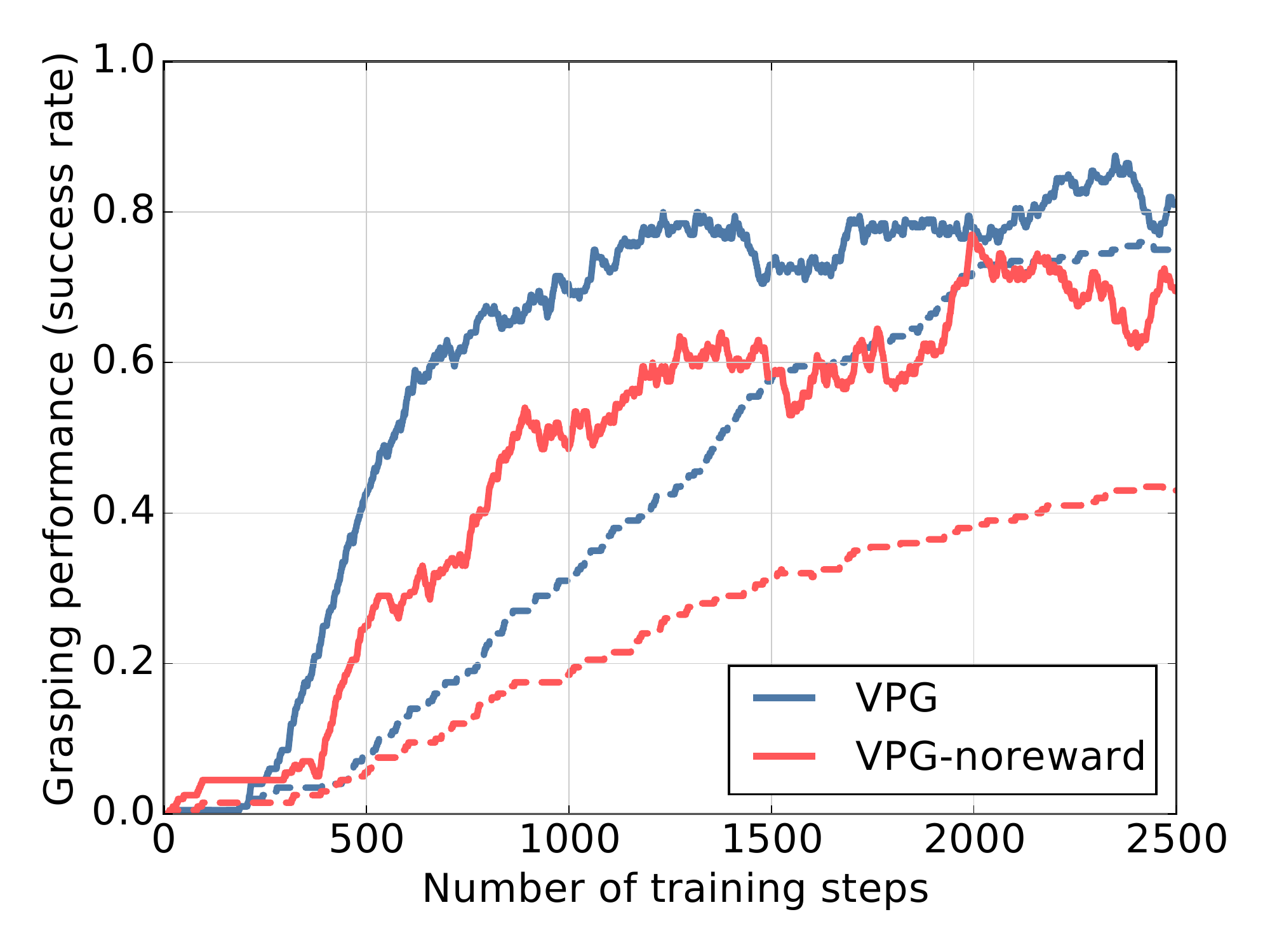}
  \vspace{-3mm}
  \caption{Comparing performance of VPG policies trained with and without rewards for pushing. Solid lines indicate \% grasp success rates (primary metric of performance) and dotted lines indicate \% push-then-grasp success rates (secondary metric to measure quality of pushes) over training steps.}
  \label{fig:train-sim-ablation}
  \vspace{-3mm}
\end{figure}

\mypara{No Pushing Rewards?}
We next investigate whether our method can learn synergistic pushing and grasping actions even without any intrinsic rewards for pushing ($R_{\mathrm{p}}(s_t,s_{t+1})=0$).  We call this variant of our algorithm ``VPG-noreward''.  In this more difficult setting, the pushing policy learns to effect change only through the reward provided by future grasps.

For this study, we run tests in simulation with 10 randomly placed objects. We report results with plots of grasping performance versus training steps.  Grasping performance is measured by the \% grasp success rate over the last $j=200$ grasp attempts, indicated by solid lines in Fig. \ref{fig:train-sim-ablation}. We also report the \% push-then-grasp success rates (\ie pushes followed immediately by a grasp -- considered successful if the grasp was successful), indicated by dotted lines. Since there is no defacto way to measure the quality of the pushing motions for how well they benefit a model-free grasping policy, this secondary metric serves as a good approximation. The numbers reported at earlier training steps (\ie iteration $i<j$) in Fig. \ref{fig:train-sim-ablation} are weighted by $\frac{i}{j}$. Each training step consists of capturing data, computing a forward pass, executing an action, backpropagating, and running a single iteration of experience replay (with another forward pass and backpropagation on a sample from the replay buffer).

From these results, we see that VPG-noreward is capable of learning effective pushing and grasping policies -- achieving grasping success rates at ~70-80\%.   We also see that it learns a pushing policy that increasingly helps grasping (note the positive slope of the dotted red line, which suggests pushes are helping future grasps more and more as the system trains). This rate of improvement is not as good as VPG, but the final performance is only slightly lower.

\begin{figure}[t]
  \centering
  \includegraphics[width=0.9\linewidth]{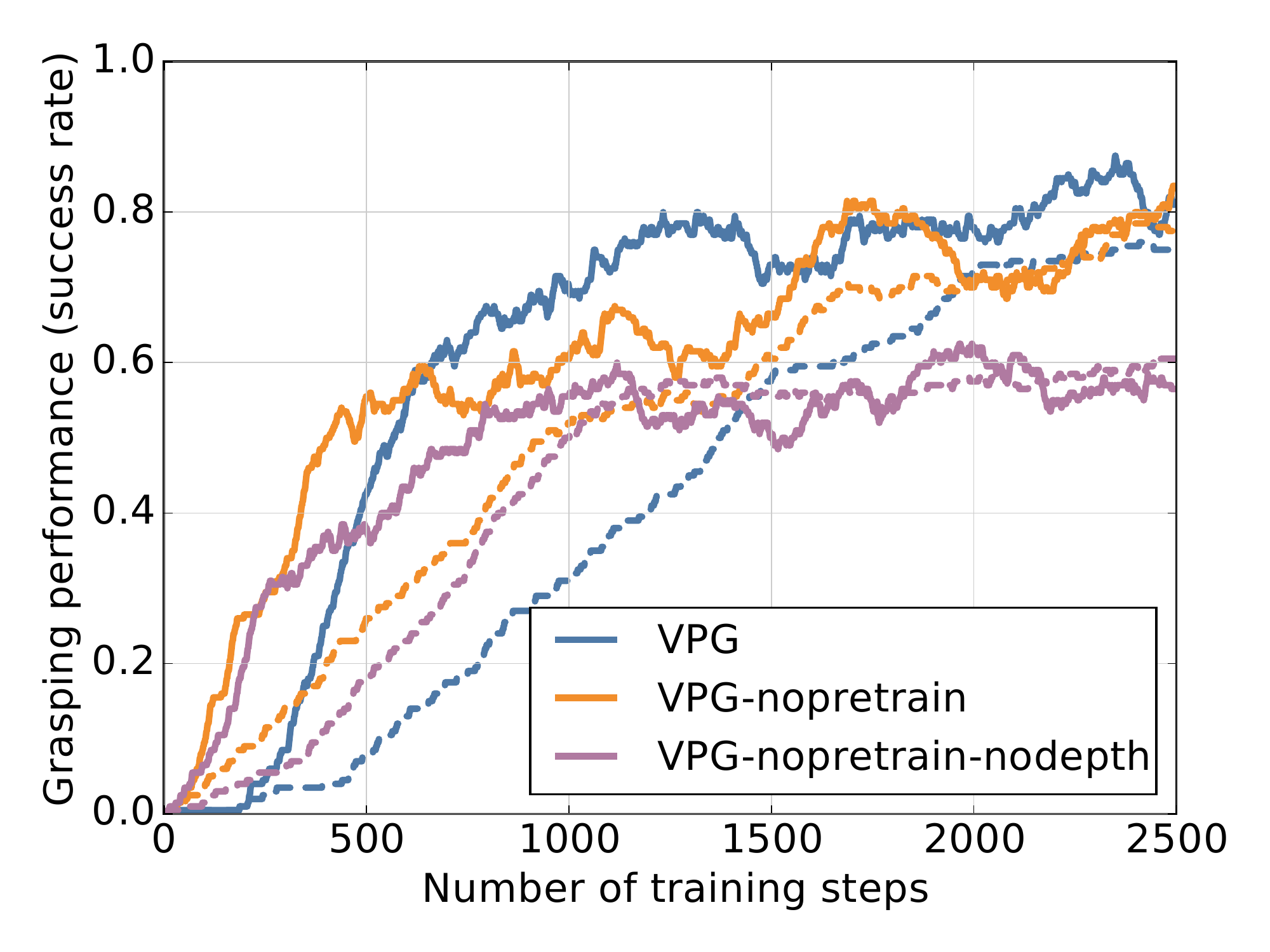}
  \vspace{-3mm}
  \caption{Comparing performance of VPG policies initialized without weights pre-trained on ImageNet and without the depth channels of the RGB-D heightmap (\ie no height-from-bottom, only color information). Solid lines indicate \% grasp success rates (primary metric of performance) and dotted lines indicate \% push-then-grasp success rates (secondary metric to measure quality of pushes) over training steps.}
  \label{fig:train-sim-ablation-appendix}
  \vspace{-3mm}
\end{figure}

\mypara{No ImageNet Pre-training?} We trained a version of VPG (``VPG-nopretrain") without ImageNet pre-training of FCN weights (\ie with only random initialization) and report its performance versus training steps in Fig. \ref{fig:train-sim-ablation-appendix}. Interestingly, the results suggest that ImageNet pre-training is not a major contributor to the sample efficiency of VPG nor to the final performance of the model in simulation. This could be due to the fact that the statistics of pixel patterns found in ImageNet images are different compared to that of re-projected heightmap images. The slight delay before the upward slope of the training curve could also be an artifact due to the FCNs spending early training steps to escape the ImageNet local optimum.

\mypara{No Height-from-bottom Information?} We trained another version of VPG (``VPG-nopretrain-nodepth") without ImageNet pre-training and without the depth channels of the RGB-D heightmap images (\ie no height-from-bottom, only color information) and report its performance in Fig. \ref{fig:train-sim-ablation-appendix}. This modification meant that each FCN $\phi_\mathrm{p}$ and $\phi_\mathrm{g}$ no longer has a second DenseNet tower to compute features from the channel-wise cloned depth channels (DDD) of the heightmaps. The results show that sample complexity remains similar, but the average final grasping performance is lower by about 15\%. This suggests that geometric cues from the depth (height-from-bottom) channels are important for achieving reasonable grasping performance with VPG. 

\mypara{Shortsighted Policies?}  
We also investigate the importance of long-term lookahead. Our Q-learning formulation in theory enables our policies to plan long-term strategies (\eg chaining multiple pushes to enable grasping, grasping to enable pushing, grasping to enable other grasps, etc.). To test the value of these strategies, we trained a shortsighted version of VPG (``VPG-myopic") where the discount factor on future rewards is smaller at $\gamma=0.2$ (trained in simulation with 10 randomly placed objects). We evaluate this policy over the 11 hard test cases in simulation and report comparisons to our method in Table \ref{table:sim-hard-ablation}.
Interestingly, we see that VPG-myopic improves its grasping performance at a faster pace early in training (presumably optimizing for short-term grasping rewards), but ultimately achieves lower average performance (\ie grasp success, action efficiency) across most hard test cases. This suggests that the ability to plan long-term strategies for sequential manipulation could benefit the overall stability and efficiency of pick-and-place.  

\begin{table}[t]
  \centering
  \setlength{\tabcolsep}{4.0 pt}
  \renewcommand{\arraystretch}{1.1}
  \caption{Comparison with Myopic Policies (Mean \%)}
  \begin{tabular}{c|c|c|c}
    \hline
    Method & Completion & Grasp Success & Action Efficiency \\\hline
    VPG-myopic & 79.1 & 74.3 & 53.7 \\
    VPG &\bf 82.7 &\bf 77.2 &\bf 60.1 \\\hline
    
  \end{tabular}
  \label{table:sim-hard-ablation}
  \vspace{-3mm}
\end{table}


\subsection{Real-World Experiments} 

\begin{figure}[h]
\centering
  \includegraphics[width=\linewidth]{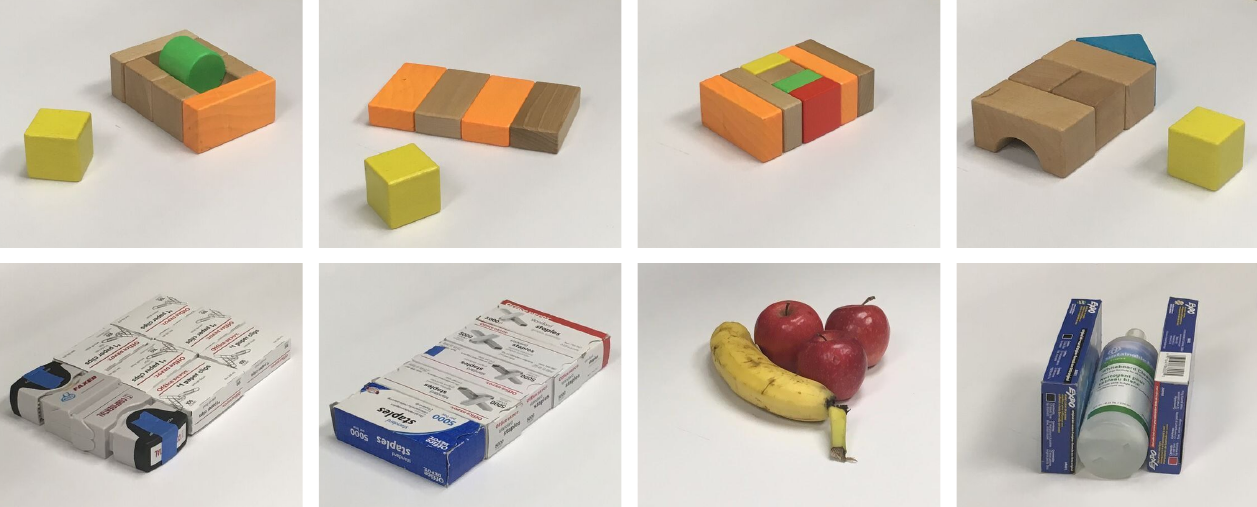}
  \caption{\textbf{Examples of challenging arrangements} in real-world settings with toy blocks (top row) and novel objects (bottom row).}
  \label{fig:test-real}
  \vspace{-3mm}
\end{figure}

In this section, we evaluate the best performing variant of VPG (with rewards and long-term planning) on a real robot. 
Our real-world setup consists of a UR5 robot arm with an RG2 gripper, overlooking a tabletop scenario. Objects vary across different experiments, including a collection of 30+ different toy blocks for training and testing, as well as a collection of other random office objects to test generalization to novel objects (see Fig. \ref{fig:test-real}). For perception data, RGB-D images of resolution $640\times480$ are captured from an Intel RealSense SR300, statically mounted on a fixed tripod overlooking the tabletop setting. The camera is localized with respect to the robot base by an automatic calibration procedure, during which the camera tracks the location of a checkerboard pattern taped onto the gripper. The calibration optimizes for extrinsics as the robot moves the gripper over a grid of 3D locations (predefined with respect to robot coordinates) within the camera field of view.

\begin{figure}[t]
  \centering
  \includegraphics[width=0.9\linewidth]{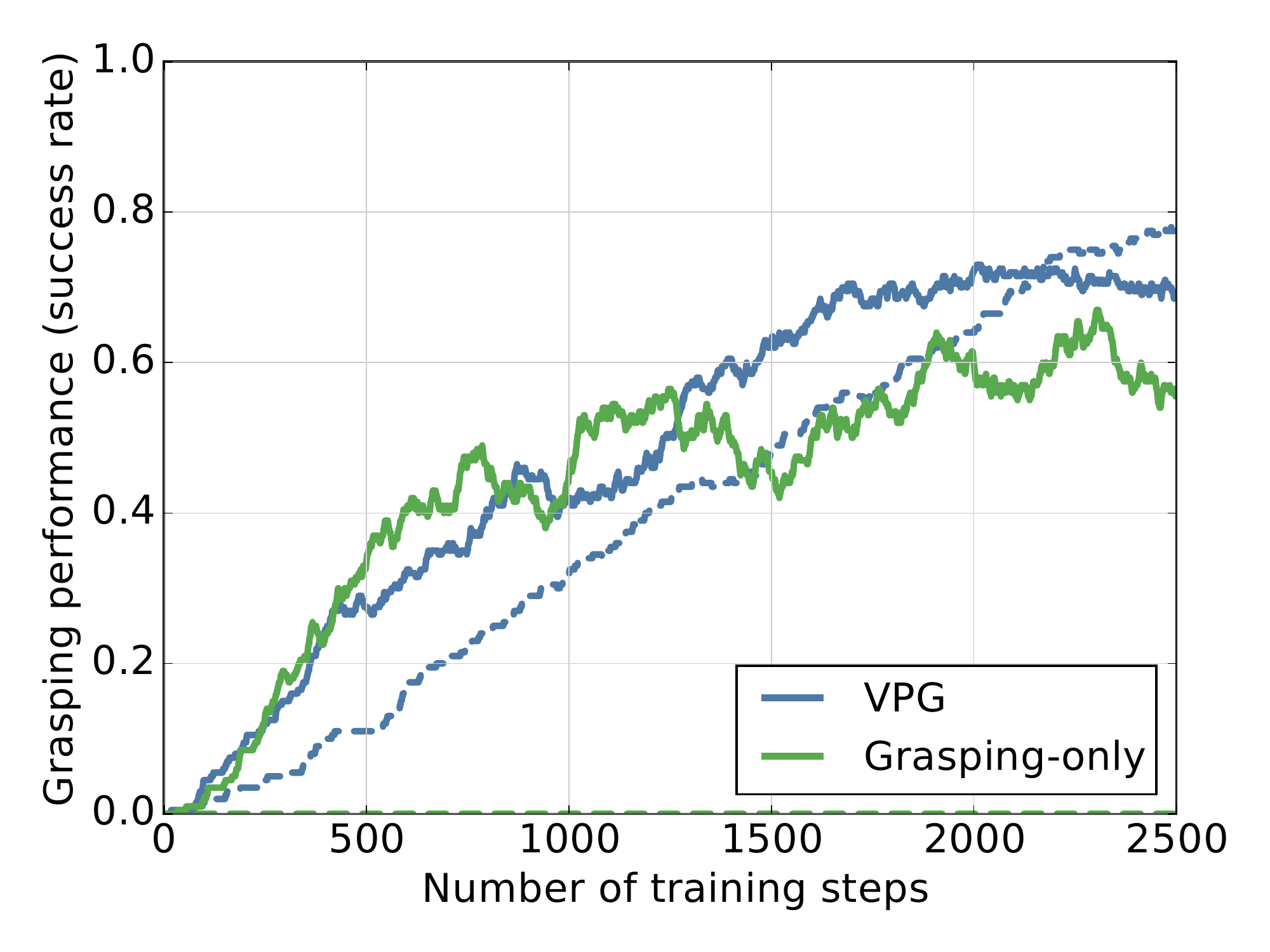}
  \vspace{-3mm}
  \caption{Evaluating VPG in real-world tests with random 30+ object arrangements. Solid lines indicate \% grasp success rates (primary metric of performance) and dotted lines indicate \% push-then-grasp success rates (secondary metric to measure quality of pushes) over training steps.}
  \label{fig:train-real-baseline}
  \vspace{-3mm}
\end{figure}

\mypara{Random Arrangements.}
We first tested VPG on the real robot in cluttered environments with 30 randomly placed objects.
Fig. \ref{fig:train-real-baseline} shows its performance versus training time in comparison to the grasping-only policy (baseline method) -- where curves show the \% grasp success rate over the last $m=200$ grasp attempts (solid lines) and \% push-then-grasp success rates (dotted lines) for both methods.

Interestingly, the improvement of performance early in training is similar between VPG and grasping-only. This is surprising, as one would expect VPG to require more training samples (and thus more training time) to achieve comparable performance, since only one action (either a grasp or a push) can be executed per training step.  This similarity in growth of performance can likely be attributed to our method optimizing the pushing policies to make grasping easier even at a very early stage of training. While the grasping-only policy is busy fine-tuning itself to detect harder grasps, VPG spends time learning pushes that can make grasping easier. 

As expected, the grasping performance of the VPG policy surpasses that of the grasping-only policy in later training steps. Not only is the performance better, it is also less erratic. This is likely because it avoids long sequences of failed grasps, which happens occasionally for grasping-only when faced with highly cluttered configurations of objects.

This experiment also suggests that VPG is quite sample efficient -- we are able to train effective pushing and grasping policies in less than 2000 transitions. At 10 seconds per action execution on a real robot, this amounts to about 5.5 hours of wall-clock training time. This is a substantial advantage over prior work on deep reinforcement learning for manipulation (\eg 10 million sample transitions (10 hours of interaction time on 16 robots) for block stacking \cite{popov2017data}).

\mypara{Challenging Arrangements.} 
We also ran experiments in the real-world comparing VPG with grasping-only on 7 challenging test cases with adversarial clutter (see examples in top row of Fig. \ref{fig:test-real}). The results appear in Table \ref{table:real-test-random}.  Note that the differences between VPG and Grasping-only are quite large in these challenging real-world cases.

\begin{table}[h]
  \centering
  \setlength{\tabcolsep}{4.0 pt}
  \renewcommand{\arraystretch}{1.1}
  \caption{Real-world Results on Challenging Arrangements (Mean \%)}
  \begin{tabular}{c|c|c|c}
    \hline
    Method & Completion & Grasp Success & Action Efficiency \\\hline
    Grasping-only \cite{zeng2017robotic} & 42.9 & 43.5 & 43.5 \\
    VPG &\bf 71.4 &\bf 83.3 &\bf 69.0 \\\hline
    
  \end{tabular}
  \label{table:real-test-random}
\end{table}

Video recordings of these experiments are provided on our project webpage \cite{projectwebpage}. They show that the VPG pushing and grasping policies perform interesting synergistic behaviors, and are more capable of efficiently completing picking tasks in cluttered scenarios in tandem than grasping-only policies.

\mypara{Novel Objects.}  Finally, we tested our VPG models (trained on toy blocks) on a collection of real-world scenes with novel objects (examples of which are shown in the bottom row of Fig. \ref{fig:test-real}). Overall, the system is capable of generalizing to sets of objects that fall within a similar shape distribution as that of the training objects, but struggles when completely new shapes or anomalies (reflective objects with no depth data) are introduced. The robot is capable of planning complex pushing motions that can de-clutter scenarios with these novel objects. We also show several video recordings of these test runs on our project webpage \cite{projectwebpage}.

\section{Discussion and Future Work}
\label{sec:discussion}

In this work, we present a framework for learning pushing and grasping policies in a mutually supportive way.  We show that the synergy between planning non-prehensile (pushing) and prehensile (grasping) actions can be learned from experience. Our method is based on a pixel-wise version of deep networks that combines deep reinforcement learning with affordance-based manipulation. Results show that our system learns to perform complex sequences of pushing and grasping on a real robot in tractable training times.

To the best of our knowledge, this work is the first to explore learning complementary pushing and grasping policies simultaneously from scratch with deep reinforcement learning. However, its limitations suggest directions for future work. First, motion primitives are defined with parameters specified on a regular grid (heightmap), which provides learning efficiency with deep networks, but limits expressiveness -- it would be interesting to explore other parameterizations that allow more expressive motions (without excessively inducing sample complexity), including more dynamic pushes, parallel rather than sequential combinations of pushing and grasping, and the use of more varied contact surfaces of the robot. A second limitation is that we train our system only with blocks and test with a limited range of other shapes (fruit, bottles, etc.) -- it would be interesting to train on larger varieties of shapes and further evaluate the generalization capabilities of the learned policies. Finally, we study only synergies between pushing and grasping, which are just two examples of the larger family of primitive manipulation actions, \eg rolling, toppling, squeezing, levering, stacking, among others -- investigating the limits of this deep reinforcement learning approach on other multi-step interactions is a significant topic for future work.

\bibliographystyle{IEEEtran} 
{\footnotesize \bibliography{main}}

\end{document}